\documentclass{article}

\usepackage{microtype}
\usepackage{graphicx}
\usepackage{subfigure}
\usepackage{booktabs} 

\usepackage{hyperref}

\usepackage[accepted]{icml2022}

\usepackage{amsmath}
\usepackage{amssymb}
\usepackage{mathtools}
\usepackage{amsthm}

\usepackage[capitalize,noabbrev]{cleveref}

\theoremstyle{plain}

\theoremstyle{definition}

\theoremstyle{remark}

\usepackage[disable,textsize=tiny]{todonotes}
\usepackage{listings}
\usepackage{xcolor}

\definecolor{codegreen}{rgb}{0,0.6,0}
\definecolor{codegray}{rgb}{0.5,0.5,0.5}
\definecolor{codepurple}{rgb}{0.58,0,0.82}
\definecolor{backcolour}{rgb}{0.95,0.95,0.92}

\lstdefinestyle{mystyle}{
    backgroundcolor=\color{backcolour},   
    commentstyle=\color{codegreen},
    keywordstyle=\color{magenta},
    numberstyle=\tiny\color{codegray},
    stringstyle=\color{codepurple},
    basicstyle=\ttfamily\footnotesize,
    breakatwhitespace=false,         
    breaklines=true,                 
    captionpos=b,                    
    keepspaces=true,                 
    showspaces=false,                
    showstringspaces=false,
    showtabs=false,                  
    tabsize=2
}

\hypersetup{
    colorlinks=true,
    linkcolor=blue,
    citecolor=blue,
    filecolor=magenta,      
    urlcolor=cyan,
    }

\icmltitlerunning{Highly Accurate FMRI ADHD Classification using time distributed multi modal 3D CNNs}

\begin{document}

\twocolumn[
\icmltitle{Highly Accurate FMRI ADHD Classification using time distributed multi modal 3D CNNs}



\icmlsetsymbol{equal}{*}

\begin{icmlauthorlist}
\icmlauthor{Christopher Sims}{Purdue}
\end{icmlauthorlist}

\icmlaffiliation{Purdue}{School of Electrical and Computer Engineering,
Purdue University, West Lafayette, IN 47906, USA}

\icmlcorrespondingauthor{Christopher Sims}{Sims58@Purdue.edu}

\icmlkeywords{Machine Learning, ICML,FMRI,3DCNN,3DGNN}

\vskip 0.3in
]



\printAffiliationsAndNotice{}  

\begin{abstract}
This work proposes an algorithm for fMRI data analysis for the classification of ADHD disorders. There have been several breakthroughs in the analysis of fMRI via 3D convolutional neural networks (CNNs). With these new techniques it is possible to preserve the 3D spatial data of fMRI data. Additionally there have been recent advances in the use of 3D generative adversarial neural networks (GANs) for the generation of normal MRI data. This work utilizes multi modal 3D CNNs with data augmentation from 3D GAN for ADHD prediction from fMRI. By leveraging a 3D-GAN it would be possible to use deepfake data to enhance the accuracy of 3D CNN classification of brain disorders. A comparison will be made between a time distributed single modal 3D CNN model for classification and the modified multi modal model with MRI data as well.
\end{abstract}

\section{Introduction}
Functional magnetic resonance imaging (fMRI) is an imaging technique that utilized MRI to measure time dependent blood flow in the brain. More specifically it measures the Blood Oxygen Level Dependent (BOLD) levels in the brain as a way to visualize the oxygen usage in different levels in the brain \cite{Heeger2002}. There are several types of fMRI techniques that are used to measure brain activity during speech, visual cues, etc \cite{LouedKhenissi2019}. However, for the diagnosis of brain related disorders such as autism, Alzheimer's disease, Bipolar disorder, and schizophrenia resting state fMRI is widely used \cite{Khosla2019}. Recently, there has been an increased interest in the use of machine learning to diagnose brain disorders from fMRI datasets  \cite{Dekhil,Khosla2019}. At the cutting edge, deep learning algorithms have taken the forefront of Attention deficit hyperactivity disorder (ADHD) diagnosis with rapid and reliable results.

Attention deficit hyperactivity disorder (ADHD) is categorized as a psychiatric disorder which is characterized my inattention, hyperactivity, etc. This condition affects about 5\%-10\% of children around the world, and has been known to persist into adulthood\cite{Polanczyk2007}. There have been three types of ADHD which have been categorized: The hyperactive-impulsive subtype (ADHD-H), The inattentive subtype (ADHD-I), and The combined hyperactive-impulsive and inattentive subtype (ADHD-C). fMRI studies have been used as a way to understand the brain activity by studying the mean activation level and comparing it to a subgroup of control subjects \cite{J.E.A.N.A.2003,Bush2005}. Recently, these datasets have become of interest in order to utilize machine learning in order to classify ADHD from fMRI data\cite{Brown}.

Resting state fMRI is a type of fMRI where the patient lays down quietly with no external stimuli and without performing any task\cite{Raichle2001,Greicius2003}. Analysis of the functional connectivity (FC) maps \cite{Bastos} of fMRI data has been successfully used to classify ADHD, the differences between subjects with ADHD and control patients has been well documented previously\cite{Cao2006,Sun2012}.

fMRI has less spatial resolution then traditional MRI to due measurement limitations, however, there is another time component that adds additional information for analysis. fMRI has played a critical in analyzing large connections between different brain regions by construction functional connectivity maps based on the collected data \cite{Smith2012}. This has resulted in the success of the human connectome project to discover connectome paths in the brain \cite{Essen2013}. The time component of fMRI can be utilized to measure abnormalities in the BOLD signal in the brain. Combined with with deep learning techniques \cite{LeCun2015}, this will allow for rapid analysis of the large data sets of fMRI data to diagnose abnormalities.

There has been a large amount of success in classifying brain disorders based of functional connectivity (FC) maps. FC maps are constructed by analyzing BOLD signals from fMRI data and constructing connectivity matrices comparing different pre-classified parts of the brain (referred to as region of interest [ROI]).\cite{Heuvel2010} From this the structural data can be extracted from fMRI data into a large set of 2D connectivity matrices that can be used as inputs into well defined 2D CNNs \cite{Yin2022}. A large drawback of utilizing this method is the need for a large amount of pre-processed of data, luckily there are databases which provide preprocessed data such as the ABIDE database \cite{Cameron}. Additionally, by constructing the FC maps, the 3D spatial data of the activity of the brain is removed the data and this takes away the ability for a deep neural network to classify a diverse set of data.

\section{Related Work} 
fMRI data has 3D spatial data with a time component to it, because of this, there has been a large interest in the utilization of 3D CNNs to analyze the spatial components without a loss of data by projecting the dataset into a 2D CNN. Within modern machine learning packages there is an ability to integrate the time component of the data. Within tensorflow \cite{Mart} this can be done with a time distributed layer.

\cite{Vu2018} developed  a 3D-CNN that used raw 3D fMRI images for disorder classification. It was proposed that this model gives a more descriptive understanding then FC based models. \cite{Zou2017} extended this to automatically diagnose ADHD by utilizing 3D-CNN on resting state fMRI datasets. Their implementation utilized commonly used 3D-CNN features in order to classify ADHD data.

\cite{Hong2021} Proposed an extension of the StyleGAN2 \cite{Karras2020}, which is a style based GAN which is able to produce highly realistic images. They presented the first implementation which is able to utilize StyleGAN2 for the generative modeling of 3D medical images (such as MRI). Their implementation is able to synthesize high quality artificial 3D medical images.

\subsection{StyleGAN2}
The latent vector $z\in Z$ is mapped by a mapping network $m$ to $w=m(z)$, $w \in W$. The synthesis network has the basis $B \in \mathbb{R}^d$ where d is the dimension. The transormed data is then modulated with the weights $A(w)\%w$ then demodulated in order to reduce artifacts in the generated data. 

\textbf{Loss Function}\\
The loss function for StyleGAN2 utilizes the logistic loss function
\begin{equation}
\mathbb{E}_{w,y \sim n(0,1)}(||\nabla_w(g(w)\cdot y)||_2 - a)^2
\end{equation}
where $w$ is the latent vector, $g$ is the generator function, and $y$ is a random image
\subsection{3D-StyleGAN}
3D styleGAN utilized a modified version of StyleGAN2 which replaces the 2D convolutions with 3D convolutions, a 3D residual network is used as the discriminator. 3D volumes need to be parsed differently in order to be integrated into StyleGAN2. Typically Learned Perceptual Image Patch Similarity (LPIPS) is used to determine the distance between an input image and the generate image $g(w,n)$ where $w$ is the latent space and $n$ is the 3D noise volume. 3D StyleGAN utilized the two mean squared error (MSE) loss at two different resolutions \cite{Hong2021} .

Since 3D images have so much data associated with them as oppose to 2D images the depth of the GAN needs to be reduced in order to same computation time and memory. It is reported that a filter depth of 128 with a latent vector size of 96 yield the best results.

\section{Problem Definition \& Methodology}
The goal of this work is to:
\vspace{-3mm}
\begin{itemize}
\item Integrate 3D StyleGAN for the MRI data part of a multi modal neural network
\vspace{-3mm}
\item implement both a single and multi modal neural network for the classification of ADHD
\vspace{-4mm}
\end{itemize}

For the fMRI data, the ADHD-200 dataset which has been pre-processed with the \href{https://www.nitrc.org/plugins/mwiki/index.php/neurobureau:AthenaPipeline}{Athena pipeline}, will be utilized. The ADHD-200 dataset contains multiple fMRI studies with different amounts of subjects. The training and validation sets will be mixed in order to mitigate errors between different datasets.

\begin{table}[h]
\caption{Datasets used from the ADHD-200 Athena pipeline}
\label{athena}
\vskip 0.15in
\begin{center}
\begin{small}
\begin{sc}
\begin{tabular}{lc}
\toprule
Dataset & \# Subjects \\
\midrule
Peking\_1    & 86 \\
KKI& 83\\
NeuroIMAGE    & 48\\
NYU   & 222\\
OHSU     & 79\\
Pittsburgh      & 32 \\
WashingtonU   & 61 \\
\bottomrule
\end{tabular}
\end{sc}
\end{small}
\end{center}
\vskip -0.1in
\end{table}

\begin{figure*}[!ht]
\vskip 0.2in
\begin{center}
\centerline{\includegraphics[width=1.0\textwidth]{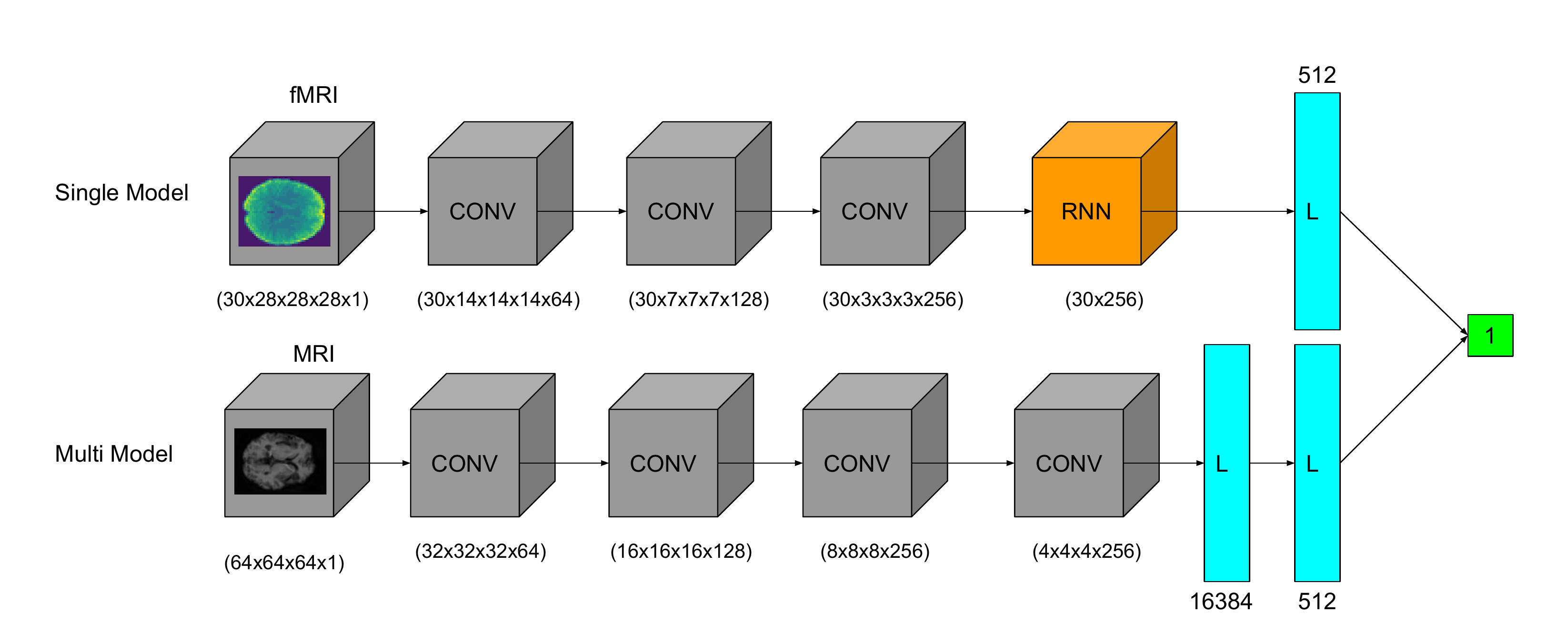}}
\caption{A Diagram of the Neural Network. CONV: 3D convolution layers, arrows indicate pooling and batch normalization steps, RNN is the recurrent neural network used for training, it can be the GRU or LSTM layer, L indicated the linear dense layers. For the multi modal model the 512 layers are concatenated into a 1024 layer. the output is a dense layer of size one for classification.}
\label{NN}
\end{center}
\vskip -0.2in
\end{figure*}

The multi modal 3D-CNN is a neural network consisting of two sub neural networks that are concatenated into one output. In previous work, complementary MRI data was feed into the neural network which is shown to have a ~3\% improvement in the classification, this work aims to integrate this methods by utilizing deep-fake data generated with a 3D-GAN. For the 3D-CNN we utilize a modified version of the method proposed by \cite{Zou2017} with time distributed layers in order to conserve the temporal component \ref{NN}. For this work, two different recurrent neural networks (RNNs) are used, Long Short Term Memory (LSTM)\cite{Hochreiter1997} and Gated Recurrent unit (GRU)\cite{Cho2014}, each network has its benefits and trade-offs. The LSTM is composed of three ``gates'' the input gate, the output gate, and the forget gate. LSTMs are a type of recurrent neural network to solve time-series data and the vanishing gradient problem. Although LSTMs are good they have an issue where there needs to be a large dataset in order to train the network. In addition, the presence of noise, LSTMs tends to shut off values due to noisy inputs \cite{Yin2017}. The GRU was developed as an iteration of the LSTM and exposes the recurrent network to the full gradient as opposed to a factor of it. In real training data, this means that the GRU tends to perform better then the LSTM in noisy and small datasets\cite{Kaiser2015}. 

\begin{center}
\textbf{Full GRU unit}
\end{center}
\vskip -0.2in
\begin{align*}
\tilde{c}_t &= \tanh(W_c [G_r * c_{t-1}, x_t ] + b_c) \\
G_u &= \sigma(W_u [ c_{t-1}, x_t ] + b_u) \\
G_r &= \sigma(W_r [ c_{t-1}, x_t ] + b_r) \\
c_t &= G_u * \tilde{c}_t + (1 - G_u) * c_{t-1} \\
a_t &= c_t
\end{align*}
\newpage
\begin{center}
\textbf{Full LSTM unit}
\end{center}
\vskip -0.2in
\begin{align*}
\tilde{c}_t &= \tanh(W_c [ a_{t-1}, x_t ] + b_c) \\
G_u &= \sigma(W_u [ a_{t-1}, x_t ] + b_u) \\
G_f &= \sigma(W_f [ a_{t-1}, x_t ] + b_f) \\
c_t &= G_u * \tilde{c}_t + G_f * c_{t-1} \\
a_t &= G_o * tanh(c_t)
\end{align*}

\subsection{3D-CNN structure}

In this paper, four neural networks are presented. With the implementation being single modality and multi modality neural networks with two different RNNs implemented

The data loader for the neural networks processes the input files in order to make them uniform throughout the dataset. the input size for fMRI is ($30\times28\times28\times28\times1$) where the last element is the number of channels for the tensor. For normal MRI images the time dimension is removed which gives a tensor input shape of ($64\times64\times64\times1$). The 3D neural network for the fMRI classifications is as follows: a time distributed layer of a 3D convolution with a kernel size of  $2\times2\times2$ which gives an output shape of ($30\times14\times14\times14\times64$) a 3D max-pooling layer with a pool size of $2\times2\times2$ and a stride of  $2\times2\times2$. Next, there is a batch normalization layer which will normalize of output for a given batch. This process is continued 2 more times for the fMRI branch and 3 times for the MRI branch of the multi modal neural network. All of these layers utilize the RELU activation function with even padding on all sides. These neural networks have filter sizes of $64$, $128$, and $256$, respectfully. Finally, the output of the last convolution layer is flattened and sent to a recurrent neural network for the fMRI data, the output is sent to a dense layer (size 512) and provided a final sigmoid activation function for classification (Figure \ref{NN}). As opposed to  previous work, which does not included time dimension as part of the 3D data, this work uses recurrent neural networks to retain information from previous time steps. There are two separate implementations with one being the LSTM RNN and the other being the GRU RNN.

For the inclusion of the MRI data, a multi modal neural network is used. A similar 3D-CNN is used for the MRI data as the fMRI layer with the only difference being that there are no time distributed layers and there are no recurrent networks. The MRI data is loaded with a separate data loader and trained with similar 3D convolutions as the fMRI model with the only difference being the lack of a time distributed augmentation. The final convultional layer gives an output of ($4\times4\times4\times256$) and is sent to a fully connected dense layer (size 512) with the sigmoid activation function as well. This dense layer is concatenated with the dense layer (512) output of the fMRI 3D-CNN network to give a new output size (1024). The final dense layer is a fully connected to a one dimensional output for classification. The goal of this implementation if to provide an ``informer'' of the brain structure for the low resolution fMRI data in order to help with the classification ADHD.

\subsection{3D-GAN MRI data generation}

\begin{figure}[!ht]
\begin{center}
\centerline{\includegraphics[width=0.9\columnwidth]{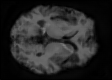}}
\caption{A Slice of a 3D MRI image generated with 3D GAN}
\label{MRI}
\end{center}
\vskip -0.2in
\end{figure}

As oppose to obtaining the MRI data associated with the patients, a randomly generated dataset of MRI images are generated with style mixing by utilizing a pre-trained 3D-GAN based on Style-GAN2. These datasets are used as a way to test to see if artificially generated data can be used to augment the training of a signal modal neural network by leveraging the geometry of a multi modal neural network. The MRI datasets (Figure \ref{MRI}) were generated with a seed vector input into the 3D GAN program.

\section{Experimental results}
In order to examine the experimental results of the proposed RNN 3D CNN neural networks, the ADHD-200 dataset was used. Data pre-processing was done by the Neuroimaging \& Resources Collaboratory (NITRC). A total of 7 different datasets were used for a total of 611 subjects. For all of the neural networks a batch size of 3 was used with a total of 10 epochs. In this work, The prediction accuracy of (typically developing control [TDC] vs. ADHD) is used as the metric for all of the neural network. The dataset was split in half with ~300 going to the training dataset and ~300 going to the validation dataset. the learning rate is set to 1$e^{-5}$ and the ADAM optimizer was utilized. Dropout was included in the RNN layers and set to a rate of 0.3. The experiment is repeated 5 times and the average result is reported herein.

\subsection{Comparison of LSTM \& GRU for single modal neural networks}
\begin{figure}[!ht]
\vskip 0.2in
\begin{center}
\centerline{\includegraphics[width=\columnwidth]{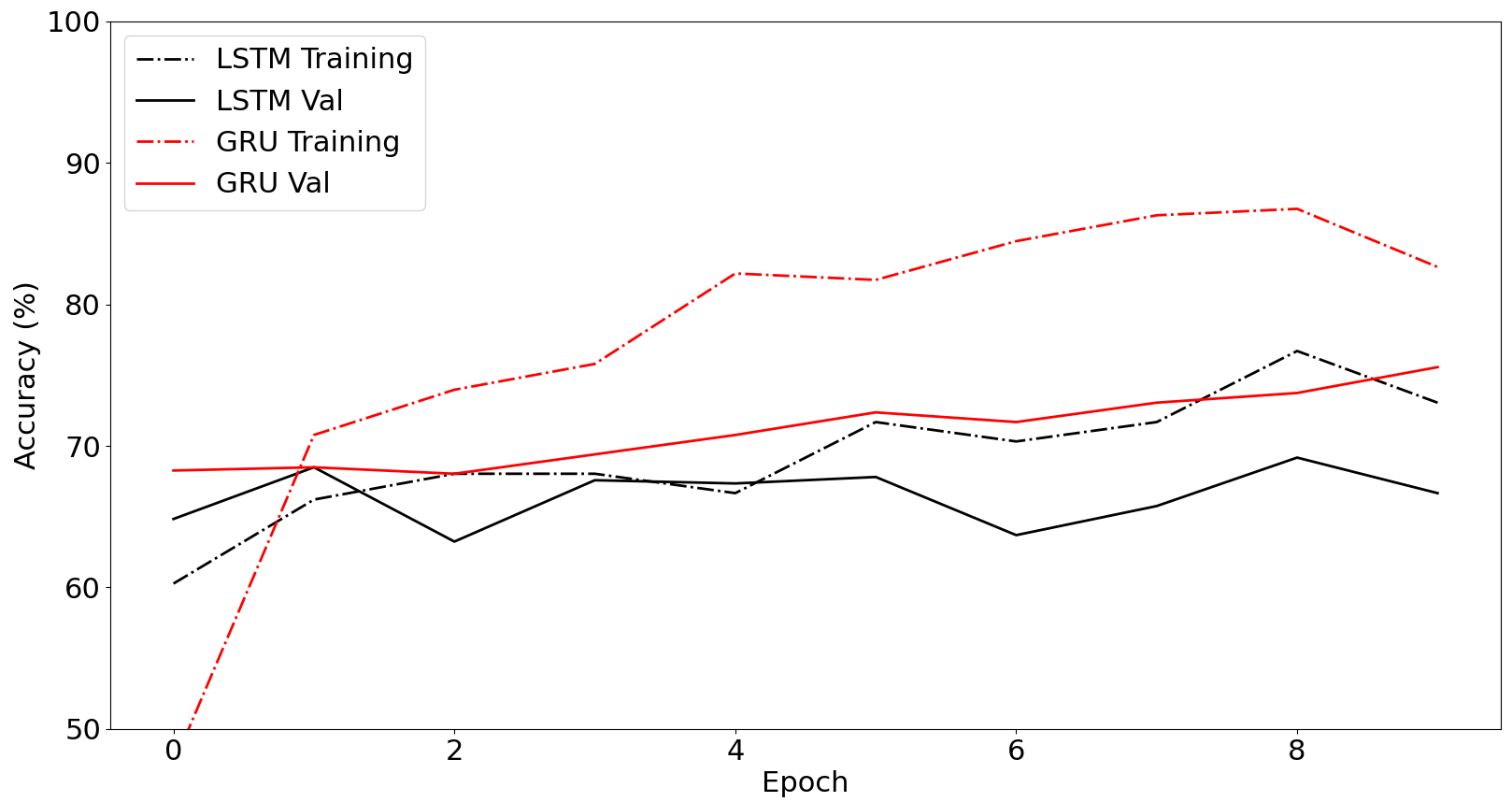}}
\caption{Training \& Validation accuracy vs Epochs}
\label{SM-ACC}
\end{center}
\vskip -0.2in
\end{figure}

The single modality RNNs show an interesting result, after the first epoch they achieve an accuracy of ~60\% and learn slowly from that value. This shows that a majority of the learning and parameters become known to the network within the first epoch. Throughout the training runs, the LSTM network stops improving its accuracy and end with an average accuracy of around 66\% (Figure \ref{SM-ACC}). 

\begin{figure}[!ht]
\vskip 0.2in
\begin{center}
\centerline{\includegraphics[width=\columnwidth]{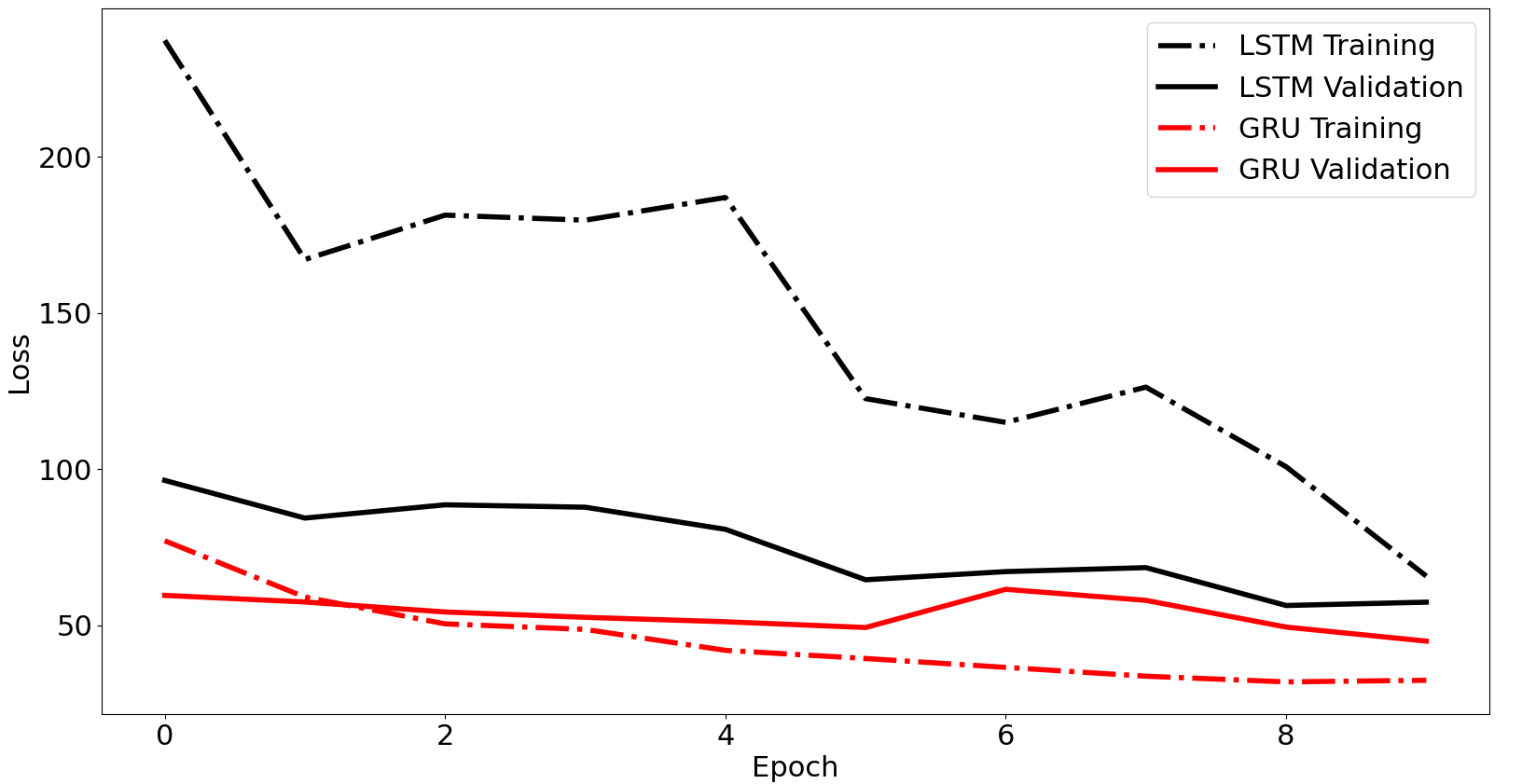}}
\caption{Loss for training the single modal network}
\label{SMLoss}
\end{center}
\vskip -0.2in
\end{figure}

However, for the GRU,  there is a remarkable improvement in the accuracy with each epoch, at the end of training the accuracy of the training dataset is around 80\% and the accuracy of the validation dataset is around 75\% (Figure \ref{SM-ACC}). This shows that, as predicted, the GRU performs better in lower quality and dataset size data. One possible explanation of this is that the GRU isn't shutting off its learning parameters as fast as the LSTM network, and the gradient is allowed to propagate trough the GRU units without being modified leading to better loss values (Figure \ref{SMLoss}).

\subsection{Comparison of LSTM \& GRU for multi modal neural networks}
Multi modal (MM) RNN networks utilize both fMRI data and MRI data in order to carry out classification in the neural network. The MRI input files for the sMRI part of the MM neural network are deepfake images generated with 3D StyleGAN. For the RNN multi modal 3D CNN, the accuracy of the training and validation has an accuracy of around 60\% after the first epoch. With more training epochs the accuracy of both the training and validation rises to 97.24\% and 97.87\%, respectively. It is found that changing the resolution of the MRI model also effects the total accuracy of the multi modal model. While an input size of ($64\times64\times64\times1$) shows high accuracy, an input size for the MRI of  ($32\times32\times32\times1$) shows a lower accuracy of around 85\% for the GRU MM model.

\begin{figure}[!ht]
\vskip 0.2in
\begin{center}
\centerline{\includegraphics[width=\columnwidth]{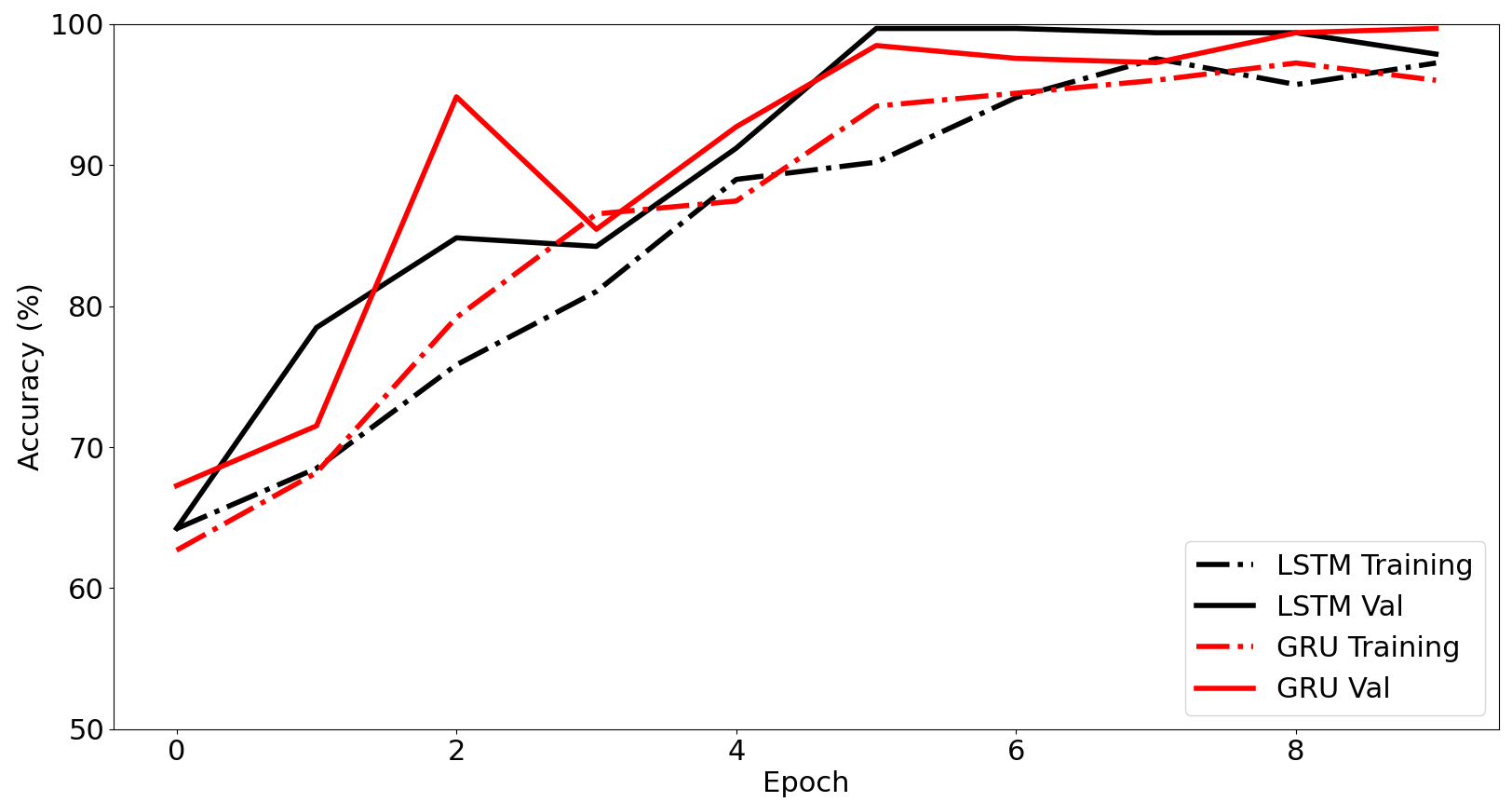}}
\caption{Training \& Validation accuracy vs Epochs for the Mutli Modal Neural Network}
\label{MM-ACC}
\end{center}
\vskip -0.2in
\end{figure}

\begin{figure}[!ht]
\vskip 0.2in
\begin{center}
\centerline{\includegraphics[width=\columnwidth]{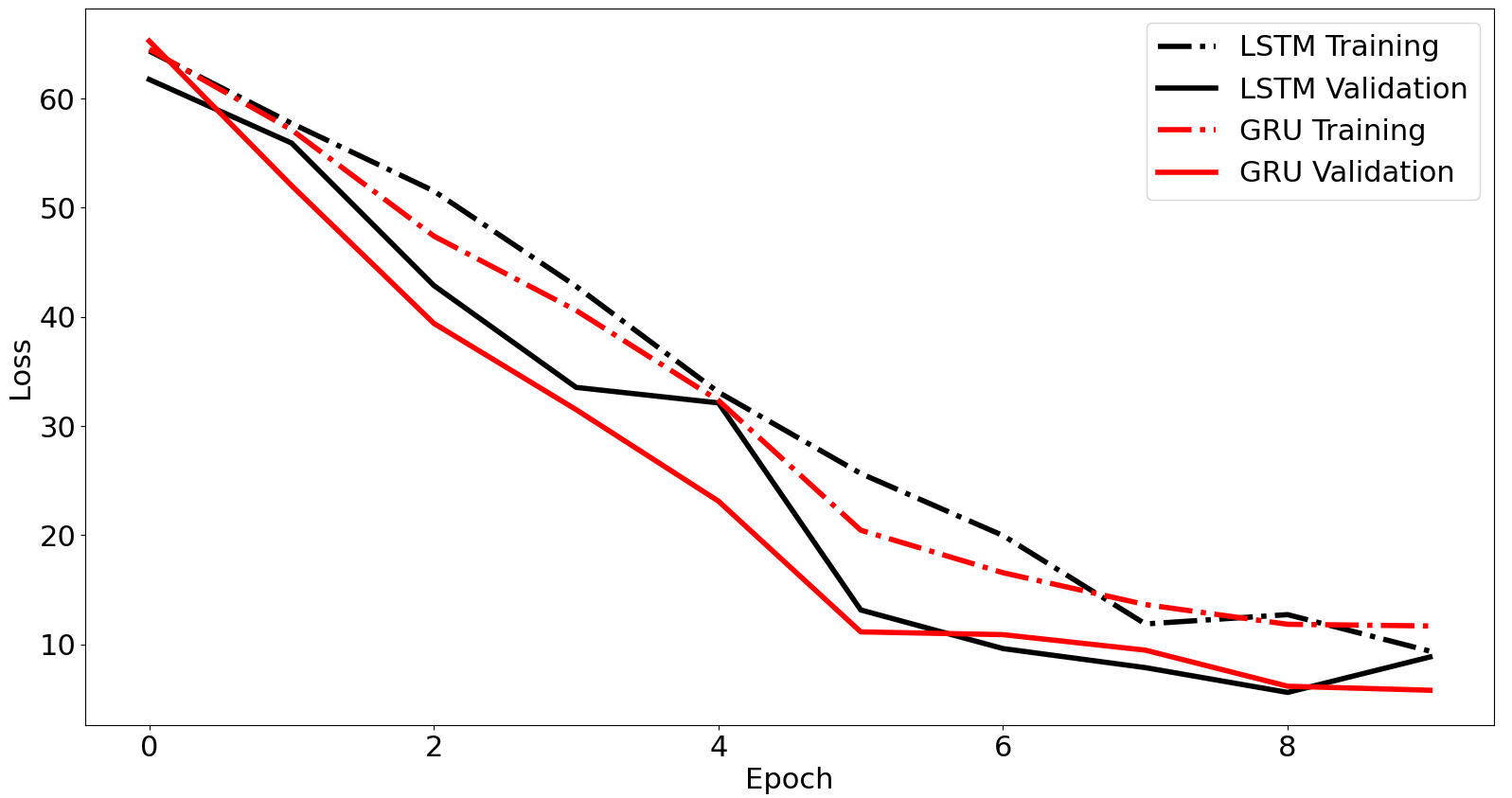}}
\caption{Loss for training the multi-modal network}
\label{MMLoss}
\end{center}
\vskip -0.2in
\end{figure}

By using a GRU the accuracy at the end of the first epoch is around 60\%, similar to the LSTM model. However, after 2 more epochs the accuracy rises rapidly to 94.84\% for the validation set. This is likely due to the random validation set having fMRI data similar to the training set. After rising, the accuracy for the validation falls to be similar to the training accuracy. The accuracy saturates above 97\% after 5 epochs, showing good convergence. By examining the loss of the same time, it is still decreasing after epoch 5. This is due to the dropout part of the network (and the GRU gates) eliminating nodes that aren't useful to the classification (Figure \ref{MMLoss}).

\subsection{Comparison with previous methods}
In order to benchmark of the neural networks presented in this work it is necessary to compare to previous works. Table \ref{methods} outlines the performance of previous works. Method 1 uses a functional connectivity map with a support vector machine (SVM)  with the multi-kernel learning  (MKL) framework utilized for the classification, method 1 reports a maximum accuracy of 61.54\%. Method 2 utilizes a machine learning model on the (2D) functional connectivity maps the the fMRI classification which achieves an accuracy of 62.57\%. Method 3 utilized a SVM and graph distances in the FC map in order to carry out classification which results in an accuracy of 62.84\%. Method 4 utilizes the social network method for classification. The social network method is a method where the edges of the neural network are defined by the correlation between two voxels, this method gives a reported accuracy of 63.75\%. Method 5 employs the most similar method to this work, a 3D neural network is carried out for classification. However, instead of retaining the temporal 4D data, method 5 uses the ReHO, VMHC or fALFF [see \cite{Zou2017} for further details] feature extraction protocols to generate a 3D dataset for analysis (without a time component). The aforementioned work also employs both single modality (66.04\%) and multi-modality 69.15\%  neural networks for classification.\\

This work utilizes a time distributed 3D neural network for classification with both a single modality and multi modality methods deployed. The single modality implementation carries out a recurrent neural network for the time evolution part of the 3D CNN. This ensures that time evolution data is well persevered throughout the training. The RNNs implemented are the LSTM block with an accuracy of 66.66\% and a GRU block with an accuracy of 75.55\%. For the multi-modal model implementation of the 3D Neural Network the LSTM network sees an average accuracy of 97.87\% for the validation, while the GRU model sees an accuracy of 99.69\%. The high performance of the 3D-GAN informed multi modal neural network warrants further discussions. As described in previous work, informing the neural network the structure of the brain has the benefit of there being a high resolution counterpart to associate data with, in addition, the minimization function will train on both the temporal fMRI data and the standard 3D image of the brain. This means that time distributed structure of the fMRI will be associated with the associated features in the brain. As opposed to utilize MRI images from the same subjects, randomly generated 3D MRI images are used. The high accuracy of this implementation shows that it is the general 3D structure of the brain that is important, and including the same MRI-fMRI dataset for each subject may cause the 3D-CNN to over train of each subjects specific brain structure as opposed to the general structure of the brain which can vary greatly form subject to subject. This general brain structure is understood the same way by humans, no matter the size of certain features of the human, we can associate the MRI/fMRI images with certain regions of the brain.

\begin{table}[h]
\caption{Previous techniques on the ADHD-200 dataset}
\label{methods}
\vskip 0.15in
\begin{center}
\begin{small}
\begin{sc}
\begin{tabular}{c c c c}
\toprule
\# & Method & Classifier & Accuracy \\
\midrule
1 & \cite{Dai} & SVM  & 61.54\%  \\
\midrule
2 & \cite{Eslami} & SVM & 62.57\%  \\
\midrule
3 & \cite{Dey} & SVM & 62.84\%  \\
\midrule
4 & \cite{Guo2014} & FC-SVM & 63.75\% \\
\midrule
		& & SM-3DCNN & 66.04\% \\
5 & \cite{Zou2017}& SM-3DCNN & 65.86\% \\
		& & MM-3DCNN & 69.15\% \\
\midrule
    &                  & SM-3DLSTM& 66.66\%  \\
6  & This work & SM-3DGRU& 75.55\%  \\
    &   		     & MM-3DLSTM& 97.87\%  \\
    & 		     & MM-3DGRU& 99.69\%  \\

\bottomrule
\end{tabular}
\end{sc}
\end{small}
\end{center}
\vskip -0.1in
\end{table}

\section{Conclusion and Future Directions}
In conclusion, recurrent neural networks (RNN) can be utilized in order to conserve the temporal part of time dependent data such as fMRI data. In doing so, more meaningful data is retained by the neural network layers leading to a better performance for the RNN, especially for the GRU which has been proven to have good performance with small datasets. This work report an average accuracy with time distributed single modal (SM) fMRI models to be 66.66\% for the LSTM based model and 75.55\% for the GRU based model.

Secondly, multi modal neural networks which are informed with 3D GAN generated MRI images greatly increase the accuracy of ADHD classification to greater then 95\%. The average accuracy of the LSTM based MM model is 97.87\% and the average accuracy for the GRU model is 99.69\%. This high performance motivates the study of multi modal neural networks of other associations in the human body. Such as heart issues measured with an electrocardiogram paired with lung structure obtained via CT-scan.

Thirdly, It is found that changing the input size can change the accuracy of the classifier used in this work. The input size must be reduced in order to reduce the number of parameters and prevent out of memory errors. A large amount of training parameters, which when paired with the large dataset can lead to this issue. This can be mitigated with the aforementioned transformer and can allow for the use of the full resolution of the dataset. All three of these issues provide a strong motivation to further study the use of transformers for classifying ADHD via fMRI data. 

Transformers with encoder-decoder architecture can be utilized due to the fact that they decease the amount of computational fact due to the fact that the data can be split and trained separately via encoding. This will allow for the full resolution of the fMRI data to be utilized (since it was down-sampled in this work). In addition, higher resolution deepfake MRI data can be used, further increasing the performance of this model.

Finally, due to the high accuracy of this model, it would prove prudent in order to append a regression, segmentation, or explainable AI model to this network in order to gain a deeper insight into what parts of the brain contribute to ADHD. Understanding which parts of the brain that are associated with brain disorders, to a high degree of accuracy, can allow for future treatment and prevention of these disorders.
In future work, developing a time-distributed 3D generative adversarial neural network would prove useful in generating deep-fake data that can be used in the prediction of ADHD via single modal 3D-CNN layers. A large issue in the field of fMRI classification is the lack of data that can be used in the training of the 3D-CNN. Augmenting the dataset with generated data could prove useful in solving the dataset issue.

\section*{Software and Data}
The code used in this work is available on \href{https://github.com/ChristopherSims/FMRI_ADHD_Classification}{GitHub}

3D StyleGAN can be found on \href{https://github.com/sh4174/3DStyleGAN}{GitHub}

StyleGAN2 can be found on \href{https://github.com/NVlabs/stylegan2}{GitHub}

The ADHD200 data set that was utilized in this paper is freely available (\href{https://www.nitrc.org/frs/?group_id=383}{NITRC})


\bibliographystyle{icml2022}

%
%

%
\end{document}